\documentclass[letterpaper, 10pt, journal, twoside]{ieeetran}
\IEEEoverridecommandlockouts

\usepackage{xparse}
\usepackage[utf8]{inputenc} 
\usepackage[T1]{fontenc}    
\usepackage{hyperref}       
\usepackage{url}            
\usepackage{booktabs}       
\usepackage{nicefrac}       
\usepackage{xcolor}         
\usepackage{bm}
\usepackage[noadjust]{cite}
\usepackage{graphicx}
\graphicspath{{figures/}}
\usepackage{hyperref}
\hypersetup{colorlinks,linkcolor={blue},citecolor={green},urlcolor={blue}}  
\usepackage{caption}
\usepackage{multirow}
\usepackage[flushleft]{threeparttable}
\usepackage{comment}
\usepackage{amsmath}
\usepackage{amssymb}
\usepackage[font=small,skip=5pt]{caption}
\usepackage[font=small,position=b,skip=5pt]{subcaption}
\usepackage[capitalise]{cleveref}
\usepackage{leftidx}
\usepackage{soul}

\NewDocumentCommand\Real{}{ \mathbb{R} }
\NewDocumentCommand\LieGroupSE{m}{ \mathrm{SE}(#1) }
\NewDocumentCommand\MatExp{m}{\mathrm{Exp}\left({#1}\right)}

\newcommand{\icratitle}{On the Coupling of Depth and Egomotion Networks for Self-Supervised Structure from Motion}
\newcommand{\shorttitle}{\icratitle}

\title{\icratitle}

\author{Brandon Wagstaff$^{1}$, Valentin Peretroukhin$^{2}$, and Jonathan Kelly$^{1\dagger}$%
\thanks{Manuscript received: January 14, 2022; Revised April 7, 2022; Accepted May 8, 2022.}
\thanks{This paper was recommended for publication by Editor Cesar Cadena Lerma upon evaluation of the Associate Editor and Reviewers' comments. 
	This work was supported in part by the Canada Research Chairs program.} 
\thanks{$^1$Brandon Wagstaff and Jonathan Kelly are with the Space \& Terrestrial Autonomous Robotic Systems (STARS) Laboratory at the University of Toronto Institute for Aerospace Studies (UTIAS), Toronto, Ontario, Canada, M3H~5T6.}
\thanks{$^2$Valentin Peretroukhin is with the Computer Science and Artificial Intelligence Laboratory at the Massachusetts Institute of Technology.}
\thanks{$^\dagger$Jonathan Kelly is a Vector Institute Faculty Affiliate.}
\thanks{\tt\footnotesize <firstname>.<lastname>@robotics.utias.utoronto.ca}%
\thanks{Digital Object Identifier (DOI): see top of this page.}
}

\begin{document}
	
\markboth{IEEE Robotics and Automation Letters. Preprint Version. Accepted May, 2022}
{Wagstaff \MakeLowercase{\textit{et al.}}: \shorttitle}	
	
\maketitle

\begin{abstract}
Structure from motion (SfM) has recently been formulated as a self-supervised learning problem, where neural network models of depth and egomotion are learned jointly through view synthesis. Herein, we address the open problem of how to best \textit{couple}, or link, the depth and egomotion network components, so that information such as a common scale factor can be shared between the networks. Towards this end, we introduce several notions of coupling, categorize existing approaches, and present a novel \textit{tightly-coupled} approach that leverages the interdependence of depth and egomotion at training time \textit{and} at test time. Our approach uses iterative view synthesis to recursively update the egomotion network input, permitting contextual information to be passed between the components. We demonstrate through substantial experiments that our approach promotes consistency between the depth and egomotion predictions at test time, improves generalization, and leads to state-of-the-art accuracy on indoor and outdoor depth and egomotion evaluation benchmarks.
\end{abstract}


\section{Introduction}

\IEEEPARstart{S}{tructure} from motion (SfM), the recovery of 3D scene structure and camera motion from a set of (monocular) images, is fundamental to vision-based state estimation for mobile robotics applications. Recently, learning-based SfM solutions have emerged that attempt to improve upon the overall accuracy and robustness of conventional SfM methods. Generally, these data-driven techniques apply convolutional neural networks (CNNs) to parameterize a direct mapping from pixel space to scene depth and camera motion. To train these networks, a self-supervised loss formulation \cite{zhou:2017} has become increasingly popular, since it obviates the requirement for ground truth labels and can facilitate online adaptation. The self-supervised training signal is generated by minimizing the photometric difference between a given `target' image and a virtual image synthesized by warping a nearby source view using the estimated scene depth and the relative camera pose (egomotion). 

Although this self-supervised approach intimately ties depth and egomotion estimation together, parameterizing both tasks through a single neural network has been shown to result in inferior performance \cite{godard2019digging}. Instead, two independent networks must be trained in such a way that their predictions are consistent in order for the photometric reconstruction loss to be minimized. In particular, these networks must implicitly adopt a mutually consistent scale factor by relying on heuristics (e.g., using the sizes of known objects within the scene) \cite{Wang_2021_ICCV}. Alternatively, it is possible to \textit{couple} (i.e., link) the depth and egomotion networks so that there is an explicit exchange of information which, for example, allows for scale details to be passed from one network to the other.

\begin{figure}[]
	\centering
	\includegraphics[width=\columnwidth]{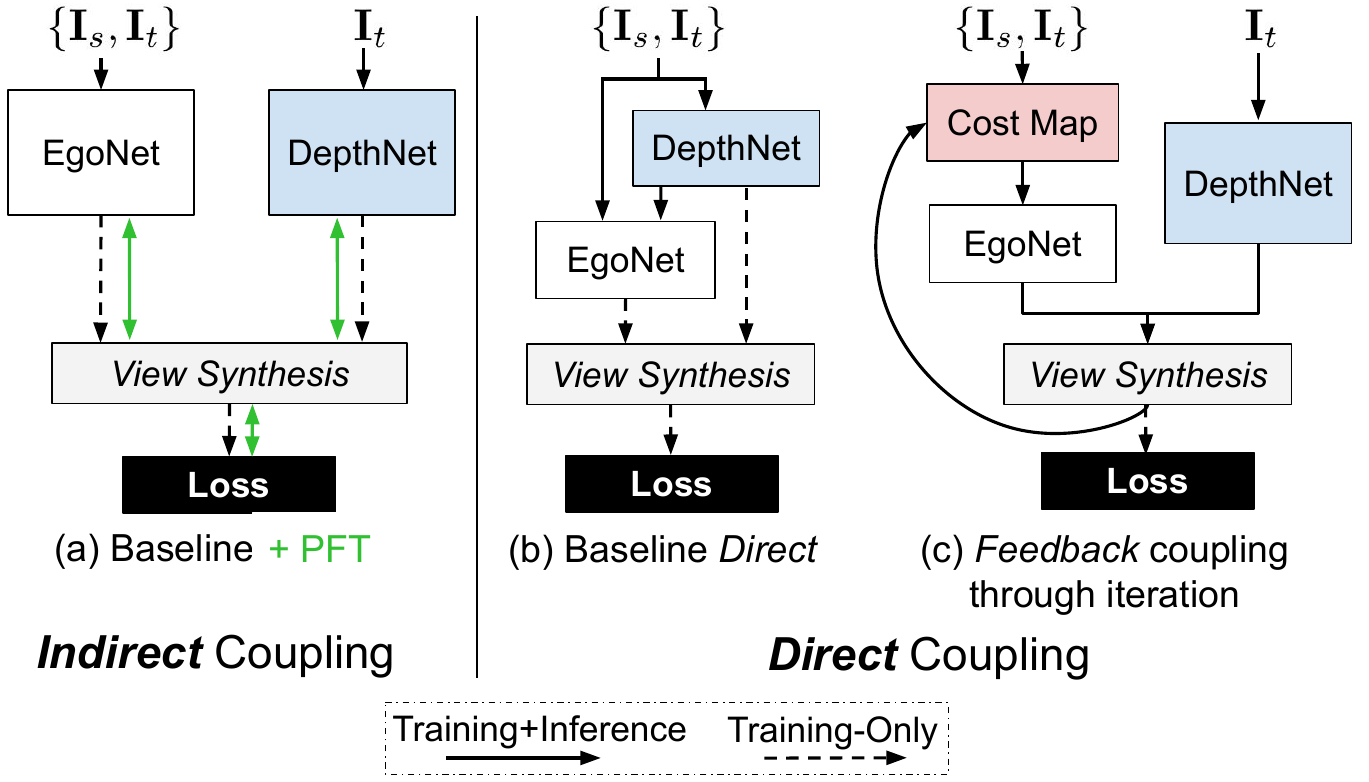}
	\caption{Three primary forms of coupling that exist between depth (DepthNet) and egomotion (EgoNet) networks. We show that feedback coupling, which in this case treats the egomotion network input as an iteratively updated cost map (i.e., an iteratively updated photo- or feature-metric input that encodes prediction error), is an effective form of coupling. In all cases, the loss is the photometric difference between a target image ($\mathbf{I}_t$) and a virtual image synthesized by warping a nearby source view ($\mathbf{I}_s$). Test-time optimization through parameter fine-tuning (PFT)	can be applied to direct coupling methods as well to retain the indirect coupling of the networks (we omit this from the diagram for simplicity).} 
	\label{fig:coupling-diagram}
	\vspace{-3mm}
\end{figure}

In this paper, we investigate and improve upon existing methods to achieve an appropriate degree of network coupling. First, we survey the field and categorize the approaches to coupling described in the existing literature (\Cref{fig:coupling-diagram,tab:coupling-methods}). We find that most systems rely on \textit{indirect coupling} of depth and egomotion via the self-supervised reconstruction loss; others incorporate \textit{direct coupling} by treating one prediction as a function of the other \cite{wang2019recurrent,li2020self}; and lastly, recent methods \cite{Nabavi2020iter,gu2021dro} incorporate a form of direct coupling that we call \textit{feedback coupling} to iteratively refine predictions based on successive forward passes. Building on this taxonomy, we present a novel network structure that ensures the depth and egomotion network predictions are \textit{tightly coupled} at both training and inference time by leveraging all three coupling strategies together. 

Our method uses iterative view synthesis \cite{Nabavi2020iter} to recursively update the egomotion network input with the most recent synthesized view, which makes the egomotion prediction a function of depth. Additionally, we incorporate test-time optimization \cite{chen2019self} for parameter fine-tuning (PFT). Through extensive experiments, we demonstrate that our unique coupling strategy promotes scale consistency between the depth and egomotion predictions, improves generalization, and leads to state-of-the-art accuracy on indoor and outdoor depth and egomotion evaluation benchmarks. Code and supplementary material are available online.\footnote{\href{https://github.com/utiasSTARS/tightly-coupled-SfM}{github.com/utiasSTARS/tightly-coupled-SfM}.}

\section{Related Work}

Early work by Zhou et al.\ \cite{zhou:2017} showed that depth and egomotion networks can be trained together by minimizing a photometric reconstruction loss. In this self-supervised training pipeline, the predicted depth and egomotion are used to differentiably warp a (nearby) source image to reconstruct the target image. Building upon \cite{zhou:2017}, recent approaches have improved the overall accuracy of the system by applying auxiliary loss terms \cite{bian2019unsupervised,mahjourian2018unsupervised,shen2019beyond}, robust feature-metric losses \cite{Zhan:2018,shu2020feature}, novel network architectures \cite{guizilini20203d,ambrus2020two,wang2019recurrent}, and techniques to mask out pixels that break the photometric consistency assumption \cite{godard2019digging,casser2019unsupervised}. In what follows, we attempt to broadly categorize the degree of coupling that exists between depth and egomotion in these systems.

\begin{table}[]
	\centering
	\caption{Recent SfM methods that use the coupling strategies shown in  \Cref{fig:coupling-diagram} between the depth (D) and egomotion (E) networks. Ours is the only method that applies all three forms of coupling.}	
	\label{tab:coupling-methods}
	\begin{threeparttable}	
		\begin{tabular}{cccc}
			\toprule
			Method & \multicolumn{3}{c}{Inference-Time Coupling Strategy} \\ \midrule
			& Indirect & Direct & Feedback \\ \midrule
			Baseline \cite{zhou:2017,godard2019digging,guizilini20203d} & --- & --- & --- \\
			PFT \cite{zhang2021deep,chen2019self,mccraith2020monocular,shu2020feature,kuznietsov2021comoda}
			& \checkmark & --- & --- \\
			Zou et al.\ \cite{zou2020learning} & --- & D$\rightarrow$E & --- \\ 
			Ambrus et al.\ \cite{ambrus2020two} & ---  & D$\rightarrow$E & --- \\ 
			Li et al.\ \cite{li2020self} & \checkmark & D$\rightarrow$E & --- \\  
			Wang et al.\ \cite{wang2020self} & --- & E$\rightarrow$D & --- \\
			ManyDepth \cite{watson2021temporal} & \checkmark & E$\rightarrow$D & --- \\
			Nabavi et al.\ \cite{Nabavi2020iter} & --- & D$\rightarrow$E & \checkmark \\
			DRO \cite{gu2021dro} & --- & D$\rightleftarrows$E & \checkmark \\
			Ours & \checkmark & D$\rightarrow$E & \checkmark \\ \bottomrule
		\end{tabular}
	\end{threeparttable}
\vspace{-3mm}
\end{table}	

\subsection{Coupling Strategies}
\subsubsection{Indirect Coupling} 

The majority of self-supervised methods \cite{zhou:2017,godard2019digging,guizilini20203d,Li:2018,bian2019unsupervised,mahjourian2018unsupervised,Vijayanarasimhan:2017} treat depth and egomotion estimation as separate tasks, and consequently employ separate networks (see Figure 1a). During training, the independently-estimated predictions are coupled as part of the view synthesis procedure, which uses depth and egomotion to reconstruct the target view from a nearby source view. We call this \textit{indirect} coupling because there is no explicit linking of depth and egomotion; rather, the weights of each network are coupled through gradient flow alone. Although this form of coupling is sufficient to jointly learn depth and egomotion, the major drawback is that the networks become decoupled at test time, which prevents contextual information (such as scale) from being passed from one network to the other. Recently, however, it has been proposed that---owing to its self-supervised nature---the reconstruction loss can be retained at inference time, and be further minimized by a gradient-descent-based optimizer \cite{li2020self,zhang2021deep,chen2019self,watson2021temporal,mccraith2020monocular,shu2020feature,kuznietsov2021comoda}. By doing so, the indirect link between networks (via gradient flow from the loss function) is preserved at inference time, and multiview geometry constraints can be enforced by minimizing the error from multiple source images. Chen et al. \cite{chen2019self} initially proposed two unique optimization approaches: parameter fine-tuning (PFT), which further optimizes the network weights, and output fine-tuning (OFT), which directly optimizes depth or egomotion predictions. Our approach uses the former for indirect coupling of the networks. 

\subsubsection{Direct Coupling} Beyond basic coupling within the loss function, other methods promote consistency between depth and egomotion by explicitly linking depth and egomotion within the network structures. Some methods \cite{wagstaff2020self1,godard2019digging} use weight sharing to effectively merge the networks; however, we note that Godard et al.\ \cite{godard2019digging} report that a baseline `shared' network structure is less accurate than a system with independent networks.
Other methods \cite{wang2019recurrent,li2020self,ambrus2020two,zou2020learning} estimate egomotion as a function of the predicted depth (see Figure 1b), with \cite{ambrus2020two} and \cite{zou2020learning} providing ablation studies indicating that doing so improves accuracy.
Less commonly, egomotion predictions have been applied to directly aid in the estimation of depth \cite{watson2021temporal,wang2020self}. 

\subsubsection{Feedback Coupling} Feedback coupling is a method that enables network \textit{introspection} by reformulating the input as a cost map built with the current depth and egomotion predictions (see Figure 1c). The cost map explicitly encodes error within the input, which the networks can utilize by iteratively updating the current prediction (i.e., through multiple forward passes). As the predictions improve, the cost map is altered, and this process is repeated until convergence (i.e., until the cost map is minimized). 
We identify two existing self-supervised feedback coupling methods in the literature. Nabavi et al.\ \cite{Nabavi2020iter} use a photometric cost map consisting of the true target view and a synthetic `target' view (generated from a nearby source view); optimal depth and egomotion predictions maximally align the synthesized view with the target, effectively minimizing the cost map. The authors of the Deep Recurrent Optimizer (DRO) \cite{gu2021dro} adopt a similar approach, but replace the photometric cost map with a feature-metric one;\footnote{Note that our definition of a cost map is generalized to encompass the variations in \cite{Nabavi2020iter} and \cite{gu2021dro}: the former implicitly encodes the error within the stacked images, while the latter explicitly computes the error within the input using the $L_2$ norm. `Minimization' in the former case refers to maximal alignment of the two images, and a zeroing of the input in the latter.} further, they employ iterative feedback for egomotion \textit{and} depth estimation, where both networks take the cost map as input. The increased complexity of this system requires a custom training procedure that alternates adjustment of the depth and egomotion network weights to maintain stability.
\begin{figure*}[t!]
	\centering
	\begin{subfigure}[]{0.55\textwidth}	
		\includegraphics[width=\textwidth]{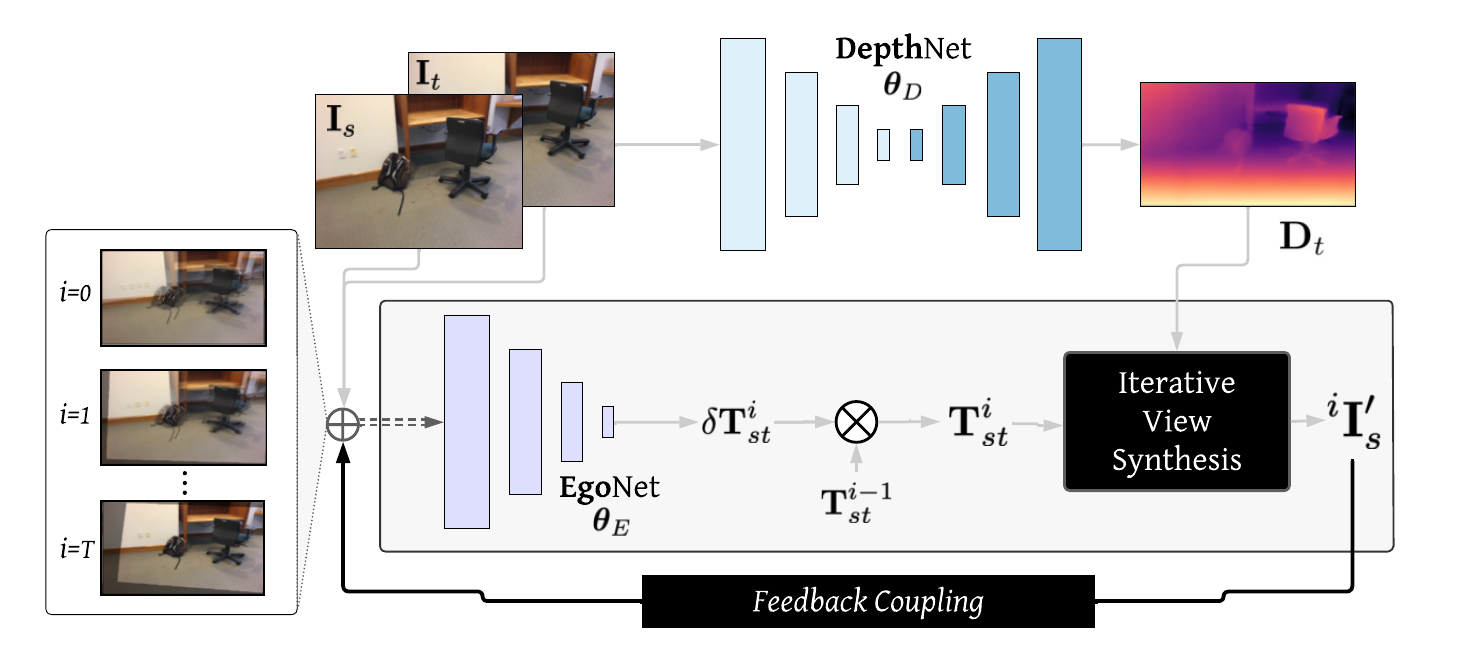}
		\caption{Our proposed network structure.}
		\label{fig:system_network}
	\end{subfigure} \hspace{0.1cm}
	\begin{subfigure}[]{0.35\textwidth}	
		\includegraphics[width=\textwidth]{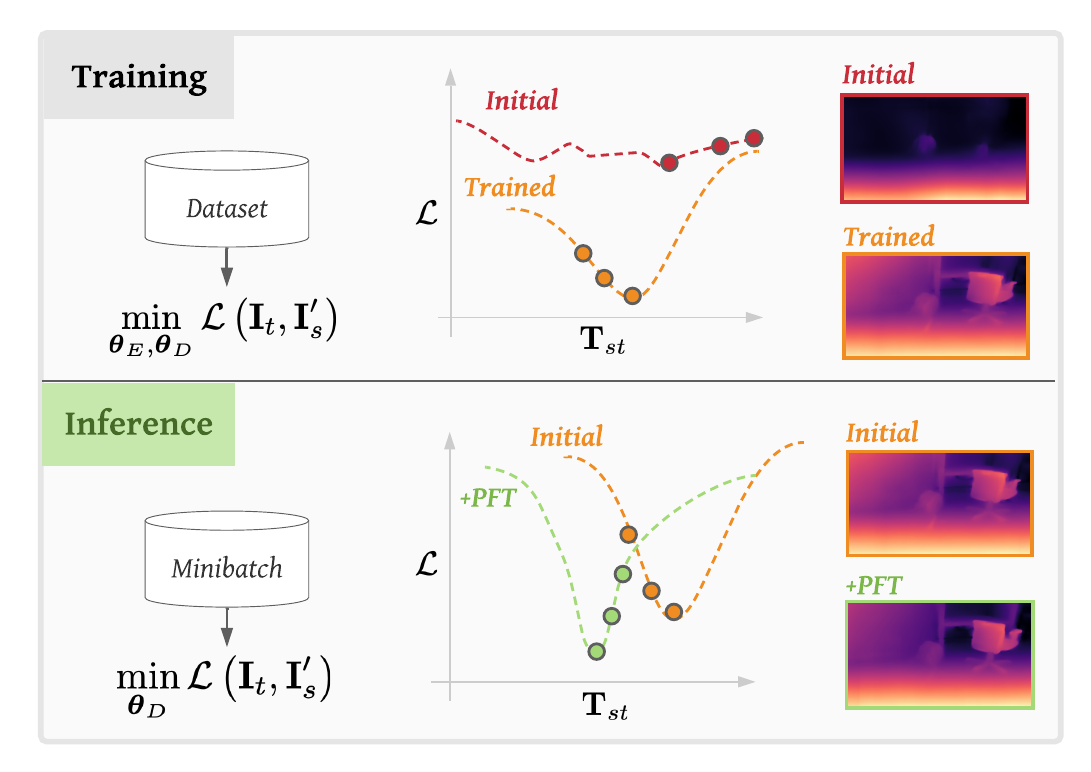}
		\caption{Our training and inference modes.}
		\label{fig:network_training}
	\end{subfigure}
	\caption{System overview. a) We use feedback coupling through iterative view synthesis \cite{Nabavi2020iter} to produce an egomotion prediction that is a function of depth. b) At inference time, the depth network weights are updated via gradient descent to further minimize the sample-wise loss. As a result of our unique coupling scheme, both depth and egomotion are improved.}
	\label{fig:system-diagram}
	\vspace{-4mm}
\end{figure*}
Finally, other methods \cite{wei2020deepsfm,tang2018ba,teed2019deepv2d,ummenhofer2017demon,clark2018learning} use iteration for feedback coupling but require supervision to train the relatively complex network structures. 

Herein, we demonstrate that feedback-based coupling is a crucial estimation component that promotes scale consistency between depth and egomotion, significantly boosting overall accuracy and improving generalization. We extend the feedback coupling approach of Nabavi et al. \cite{Nabavi2020iter} by incorporating indirect coupling into our system using an inference-time PFT strategy to achieve a \textit{tight coupling} of predictions, that is, our approach links depth to egomotion, and vice versa, such that an improvement in one leads to an improvement in the other.
\section{Approach}
We detail our tightly-coupled approach in three parts. First, we introduce our baseline (decoupled) depth and egomotion networks and the self-supervised loss formulation used for training. Second, we describe feedback coupling based on iterative view synthesis. Finally, we present our inference-time depth optimization technique that forms the final component of our framework for estimating depth and egomotion. See \Cref{fig:system-diagram} for an illustration of our system.

\subsection{Baseline Depth and Egomotion Framework}
\label{one-shot-framework}

For a single image pair consisting of a source and target image, $\mathbf{x} = \{\mathbf{I}_s, \mathbf{I}_t\}$, the latent variables of interest are the target image depth $\mathbf{D}_t \in \Real^{H\times W}$ and the inter-frame pose change (egomotion) $\mathbf{T}_{st} \in \LieGroupSE{3}$ between the images. We produce these estimates from two separate networks: $\mathbf{D}_t = f_{\bm{\theta}_D}(\mathbf{I}_t)$, $\mathbf{T}_{st} = f_{\bm{\theta}_{E}}(\mathbf{I}_s, \mathbf{I}_t)$. These networks are jointly trained by minimizing a photometric reconstruction loss $\mathcal{L}_{P}$ that compares a reconstructed target image $\mathbf{I}'_s$ with the observed target image,
\begin{equation} \label{eq:l_phot}
\mathcal{L}_{P} = (1-\alpha)\left|\mathbf{I}'_{s} - \mathbf{I}_{t} \right| + \alpha \mathcal{L}_\text{SSIM}(\mathbf{I}'_{s},\mathbf{I}_{t}),
\end{equation}
which is a weighted combination of an $L_1$ loss and a structural similarity (SSIM) loss \cite{wang2004image}. A spatial transformer \cite{Jaderberg:2015} is used to reconstruct $\mathbf{I}'_s$ from $\mathbf{I}_s$ by populating each target image pixel coordinate $\mathbf{u}'$ with the value from its corresponding location in the source image,
\begin{align}
\label{eq:loss_terms}
\mathbf{I}'_s(\mathbf{u}') = \mathbf{I}_s(\mathbf{u}), & \quad \mathbf{u} = \pi(\mathbf{T}_{st}\pi^{-1}(\mathbf{u}')).
\end{align}
A pinhole camera projection model $\pi(\mathbf{p}) = \mathbf{K}\frac{1}{z}\mathbf{p}$ relates a 3D point in the scene to its 2D image coordinates, and the inverse model $\mathbf{p} = \pi^{-1}(\mathbf{u})$ does the opposite, using the estimated per-pixel depth $z$. We include two regularization losses to encourage the depth network to produce a realistic representation of the scene: a smoothness loss $\mathcal{L}_{s}$ \cite{godard2017unsupervised}, and a geometric consistency loss $\mathcal{L}_{G}$ \cite{bian2019unsupervised}. Our total \textit{per-sample} loss is then:
\vspace{-0.2cm}
\begin{align}
\mathcal{L} = \lambda_{P}\mathcal{L}_{P}(\mathbf{I}'_s, \mathbf{I}_t) + \lambda_{S}\mathcal{L}_{S}(\mathbf{I}_t,\mathbf{D}_t) + \lambda_{G}\mathcal{L}_{G}(\mathbf{D}_t, \mathbf{D}_s, \mathbf{T}_{st}).
\end{align}

This baseline system employs no coupling aside from the indirect coupling that happens during training. To better couple our predictions, we use the feedback coupling approach below, which relies on iterative view synthesis.

\subsection{Feedback-Coupled Egomotion Prediction}
\label{sec:iterative_method}
   
\begin{figure}[b!]
	\vspace{-3mm}
	\centering
	\begin{subfigure}[]{0.49\textwidth}
		\centering
		\includegraphics[width=0.50\textwidth]{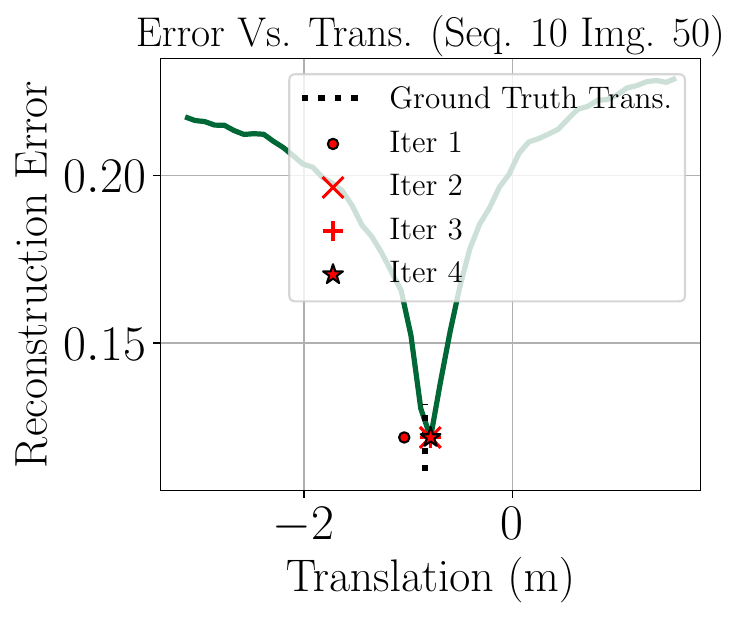}	
		\includegraphics[width=0.48\textwidth]{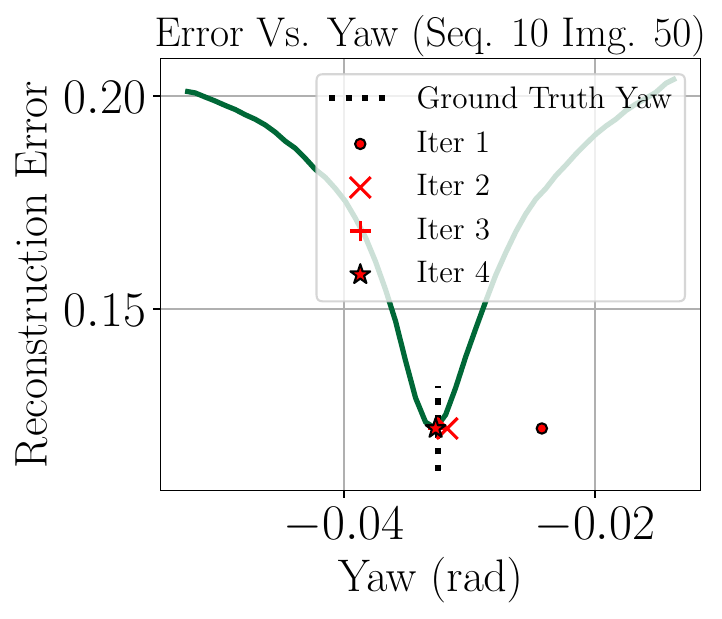}
		\caption{Feedback coupling iteratively updates the egomotion prediction to minimize the loss function.}
		\label{fig:loss_curves_no_opt}
	\vspace{1mm}
	\end{subfigure}
	\begin{subfigure}[]{0.49\textwidth}
		\centering
		\includegraphics[width=0.49\textwidth]{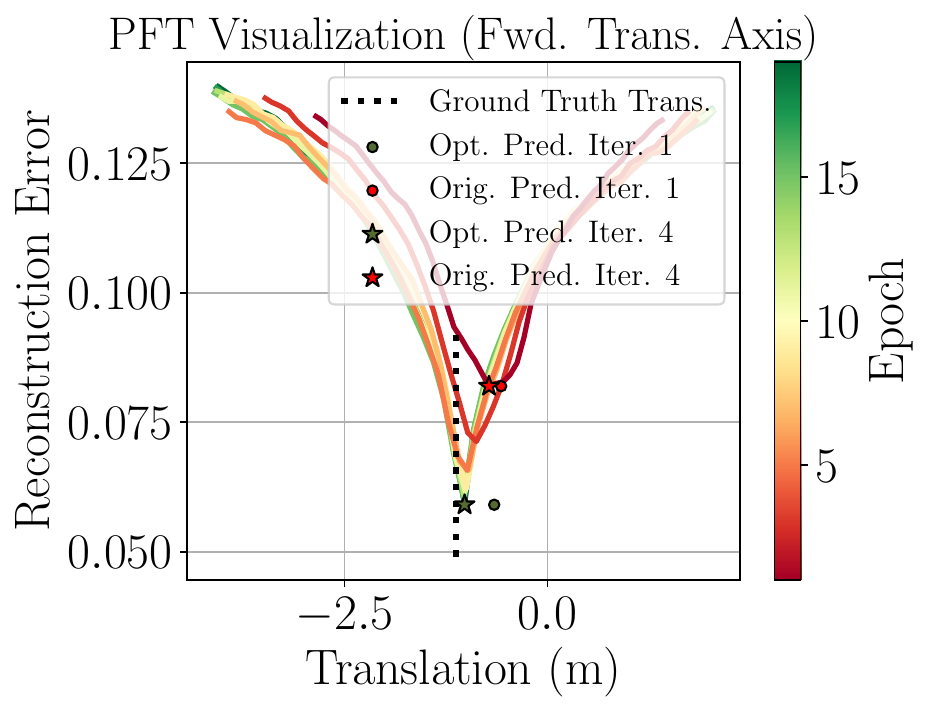} 
		\includegraphics[width=0.49\textwidth]{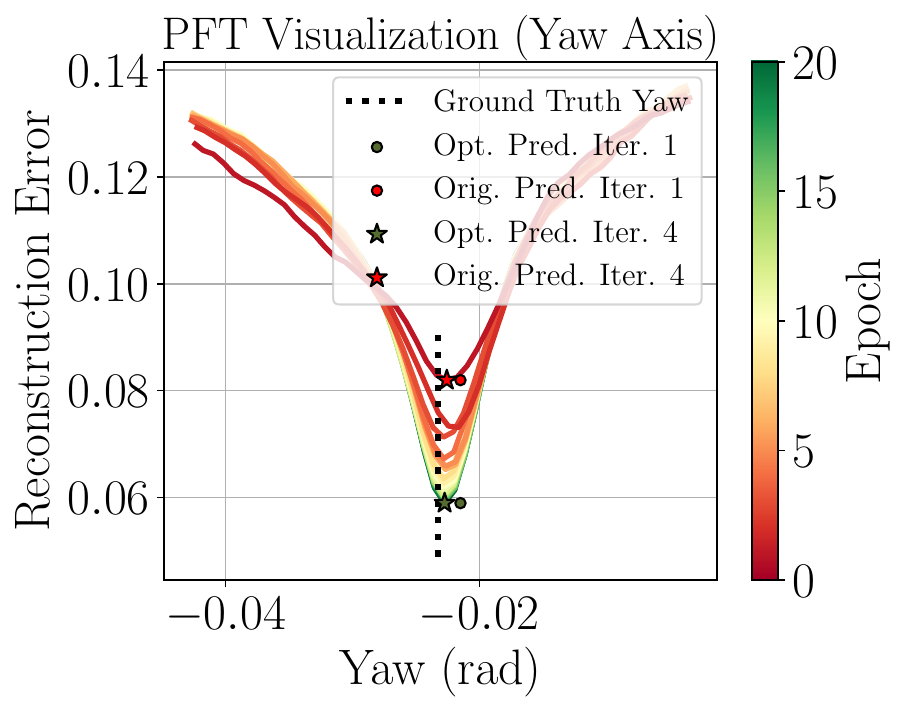} 
		\caption{Incorporating indirect coupling through PFT reduces the overall loss by improving depth accuracy.}
		\label{fig:loss_curves_opt}
	\end{subfigure}
	\caption{One-dimensional loss experiments illustrating the impact of coupling at inference time.}
\end{figure}

In line with Nabavi et al. \cite{Nabavi2020iter}, we extend the standard egomotion estimation approach to incorporate feedback through iterative view synthesis. We perform multiple forward passes through the network and redefine the egomotion network input as a photometric cost map that is updated after each iteration.
Our network structure remains the same as the baseline network from \Cref{one-shot-framework}; the only difference is that multiple passes through the network are made, where the $i^\text{th}$ pass takes, as an input, a recursively-updated cost map based on the images $\{^i\mathbf{I}'_s, \mathbf{I}_t\}$. Each pass through the network produces a correction $\delta \bm{\xi}^i_{s't} = f_{\bm{\theta}_E}(^i\mathbf{I}'_s, \mathbf{I}_t)$ that better aligns the current reconstructed source image $^i\mathbf{I}'_{s}$ with the target image. The corrections are compounded to produce the full pose change between the original source and target image,
\begin{equation}
\mathbf{T}_{st} = \left(\prod_{i=1}^{T}\delta \mathbf{T}^i_{s't}\right)\mathbf{T}^0_{st}
\approx \MatExp{\sum_{i=1}^{T}\delta \bm{\xi}^i_{s't}}\mathbf{T}^0_{st},
\end{equation} 
where we parameterize each correction as an unconstrained vector $\delta \bm{\xi}^i_{s't} \in \Real^{6}$, and then apply the (capitalized) exponential map \cite{Sola:2018} to produce an on-manifold $\LieGroupSE{3}$ correction; the summation is a reasonable approximation because the corrections are generally very small.
We view the first pass through the network to be a correction to an \textit{a priori} null egomotion initialization $\mathbf{T}^0_{st}$ that (more coarsely) aligns the original source image with the target image. 
In subsequent iterations, the warped image and the target image are used as inputs to produce an egomotion correction. Notably, since $^i\mathbf{I}'_s$ is a function of the current depth and egomotion predictions, the corrections can take into account the error that is explicitly encoded into the input cost map. Importantly, scale inconsistency between the network predictions can also be accounted for in subsequent iterations.
To demonstrate the impact of incorporating iteration, we visualize some sample-specific photometric reconstruction loss curves (as a function of egomotion) in \Cref{fig:loss_curves_no_opt}.\footnote{The 1D loss curves were generated by sampling forward translation or yaw values via a grid search and then plotting the resulting $\mathcal{L}_P$ produced using these sampled values and the current depth prediction. The egomotion network's actual prediction at each iteration is overlayed on this curve.} Here, we observe that the egomotion predictions converge to the minimum of the loss function. \Cref{fig:losscurves} illustrates how applying more iterations improves convergence and generalization during training. Note that if the input depth prediction is erroneous, the minimum loss will not always align with the ground truth value of the egomotion; we address this by refining depth predictions through PFT, which we describe next. 

\subsection{Tightly-Coupled Depth and Egomotion Optimization}
We extend the feedback coupling method linking depth with egomotion by doing the converse: linking egomotion with depth. This additional network coupling is achieved by incorporating an inference-time PFT strategy: the depth network weights are refined by further minimizing the self-supervised loss through gradient descent.
By incorporating our feedback-coupled egomotion network with our depth optimizer, we can achieve \textit{tightly-coupled} optimization of depth and egomotion. The novelty of our optimization procedure with respect to other PFT methods is that we can produce refined depth \textit{and} egomotion predictions through the optimization of our depth network only, because our (feedback-coupled) egomotion predictions are already a function of depth. As the depth predictions are refined by PFT, we recompute an improved egomotion prediction via iteration also (where the cost map input is updated with the refined depth). 

\begin{figure}[]
	\centering
	\begin{subfigure}[]{0.47\textwidth}
		\includegraphics[width=\textwidth]{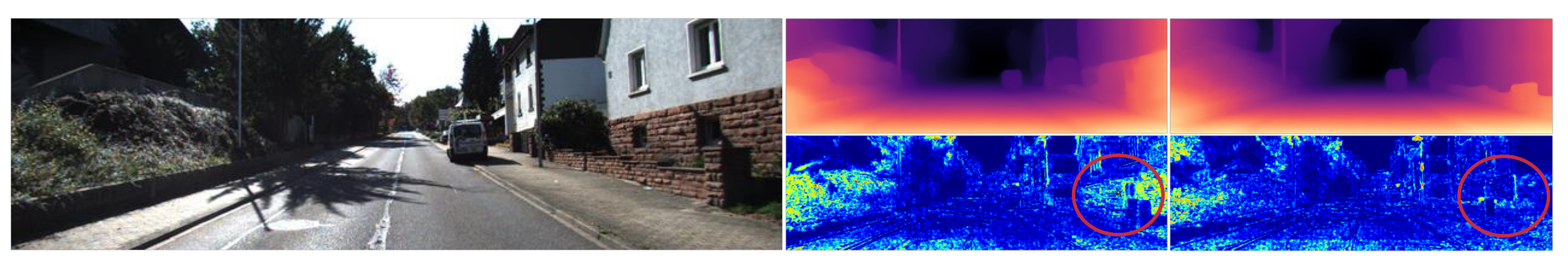}
		\caption{KITTI dataset example (Seq. 09 frame 751).}
		\label{fig:depth_opt_vis_kitti} 
	\vspace{1mm}
	\end{subfigure} 
	\begin{subfigure}[]{0.47\textwidth}
		\includegraphics[width=\textwidth]{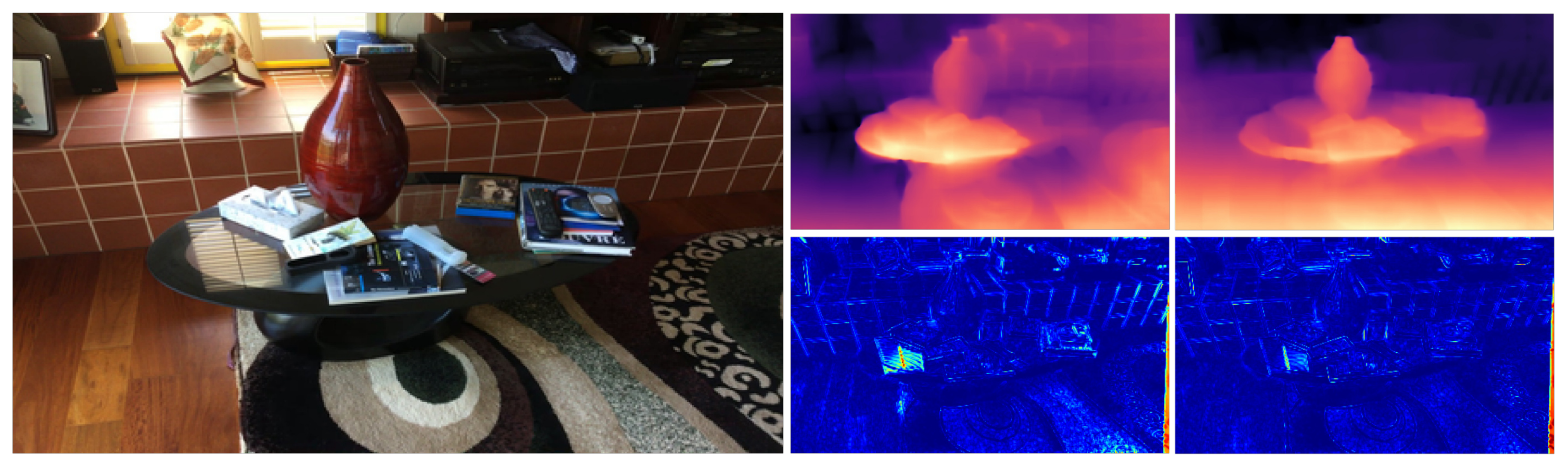}
		\caption{ScanNet dataset example (Scene0672\_01 frame 000017).}
		\label{fig:depth_opt_vis_scannet} 
	\end{subfigure}
	\caption{Depth predictions and reconstruction errors before (middle) and after (right)  inference-time depth optimization.}
	\label{fig:depth_opt_visualize}
	\vspace{-3mm}
\end{figure}

To perform PFT at inference time, we minimize the same loss as \cref{eq:loss_terms} but replace the smoothness term with a depth prior that ensures that the optimized depth $\mathbf{D}^*_t$ remains similar to the original depth prediction:
\begin{align}\label{eq:opt-loss}
\mathcal{L} = \lambda_{P}\mathcal{L}_{P} + \lambda_{G}\mathcal{L}_{G} + \lambda_{prior}\mathcal{L}_{SSIM}(\mathbf{D}^*_t, \mathbf{D}_t).
\end{align}
\Cref{fig:loss_curves_opt} illustrates our tightly-coupled optimization procedure in one dimension for a single inference-time sample. Beginning with the initial (red) loss curve produced by our trained depth network, the egomotion network recursively updates its predictions to converge to the (suboptimal) minimum. Running through each epoch of the inference-time depth optimizer shifts the loss to a new (lower) minimum by improving the quality of the depth prediction; then, given the new depth prediction, our egomotion network is able to converge to the new minimum. We visualize an example of the improvement in depth accuracy (and the corresponding reduction in the photometric reconstruction error) in \Cref{fig:depth_opt_visualize}.

\section{Experiments \& Results}

\begin{figure*}[]
	\centering
	\begin{subfigure}[]{0.65\textwidth}
		\centering
		\includegraphics[width=0.46\textwidth]{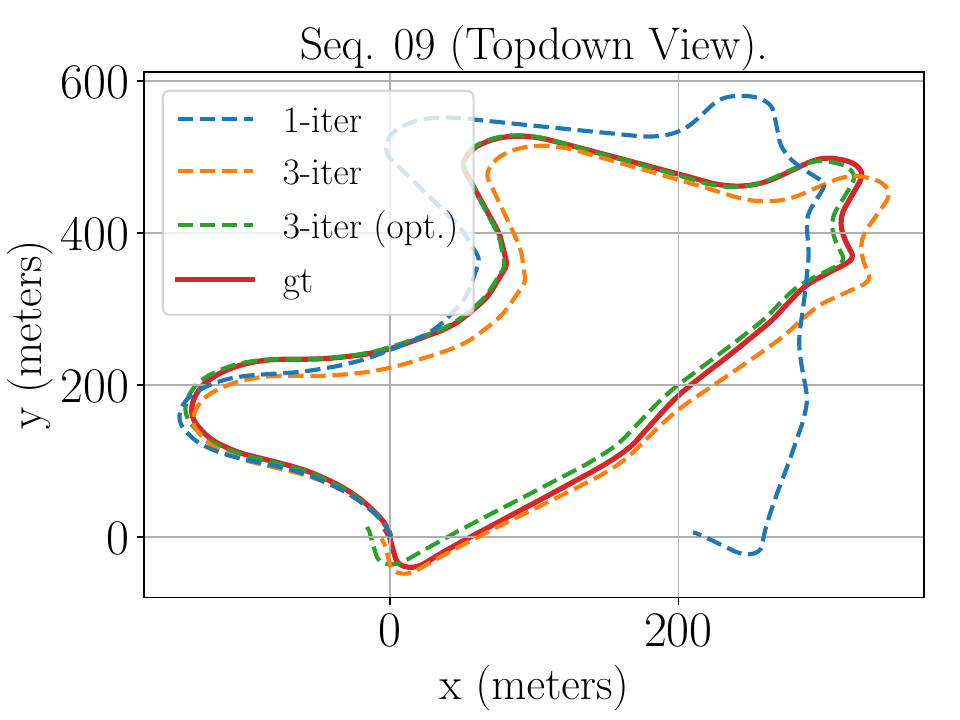} 
		\includegraphics[width=0.46\textwidth]{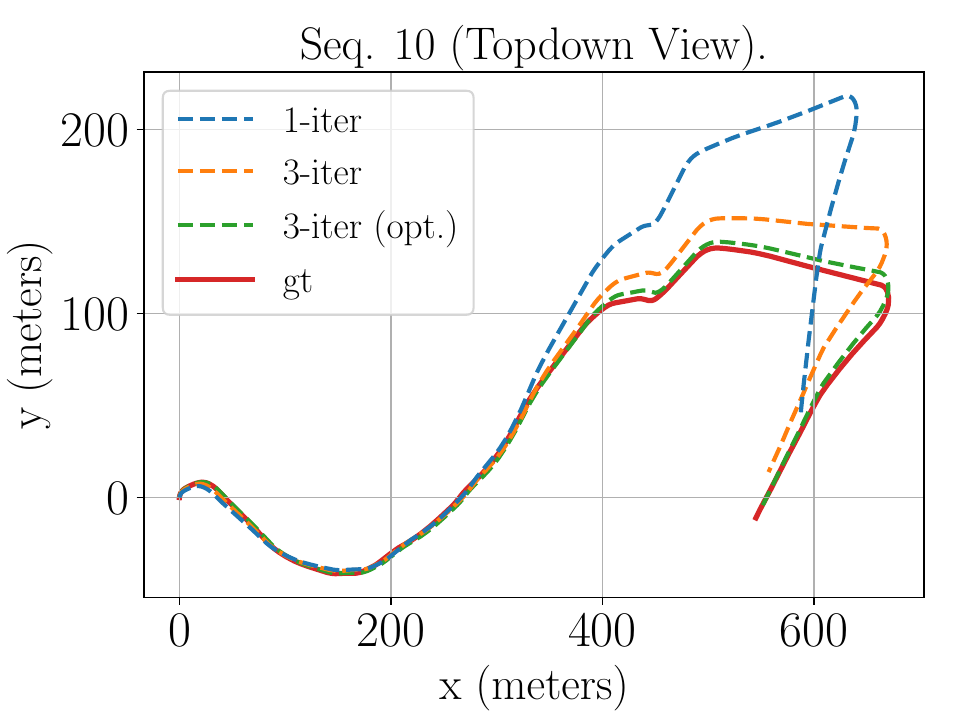}
		\vspace*{-1mm}
		\caption{KITTI odometry results on the test sequences.}
		\label{fig:topdown_traj_kitti}
	\end{subfigure}
	\begin{subfigure}[]{0.32\textwidth}
		\includegraphics[width=\textwidth]{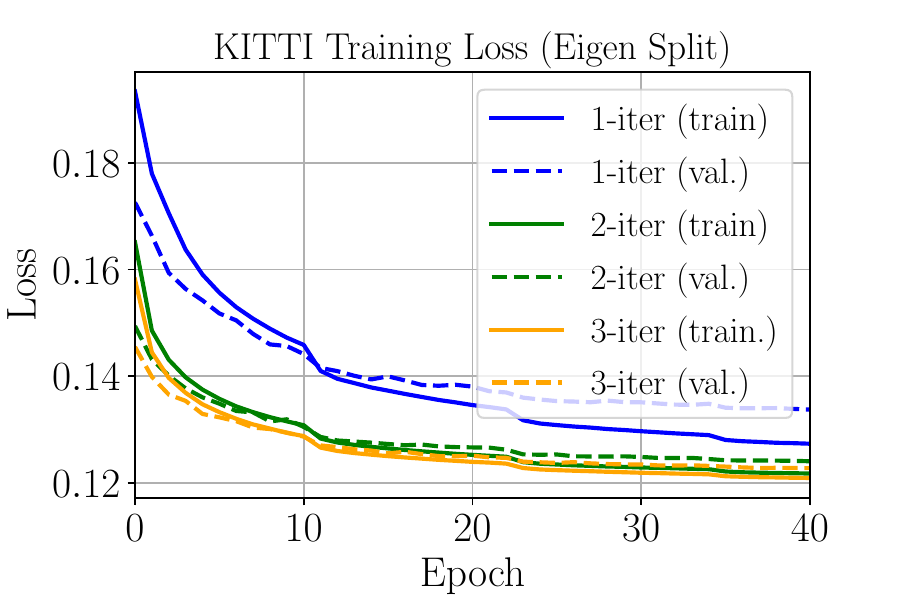}
		\caption{Training losses for the Eigen split.}
		\label{fig:losscurves}
	\end{subfigure}
	\caption{Results for the KITTI dataset. Our tightly-coupled system leads to (a) improved egomotion accuracy, and (b) improved convergence and generalization compared to a baseline (one-iteration) method that does not employ inference-time network coupling.}
	\label{fig:kitti_results}
	\vspace{-4mm}
\end{figure*}

We provide details of our network structure and training procedure below, followed by extensive experimental results on the ScanNet, KITTI, and Oxford RobotCar datasets. Specifically, we evaluate the performance of our tightly-coupled system on depth and egomotion benchmarks from KITTI and ScanNet, showing that we achieve state-of-the-art accuracy on the KITTI odometry benchmarks and the ScanNet benchmarks. Further, we include ablation studies that indicate how both feedback and indirect coupling, although useful on their own, complement each other when combined. Finally, we demonstrate additional benefits of feedback coupling. In particular, this added coupling improves generalization across datasets, which we demonstrate through cross-dataset evaluation (from KITTI to Oxford RobotCar). Additional information is available in our supplementary material.
\subsection{Datasets} ScanNet \cite{dai2017scannet} is a large indoor dataset with 2.5 M views in 1,513 sequences. We follow the training/test split from \cite{tang2018ba}, where the first 1,413 sequences are used for training, and 2,000 image pairs from the remaining 100 sequences are selected for testing. Our networks use images downsized to 256 $\times$ 448 pixels. The KITTI dataset \cite{Geiger:2013} is an outdoor driving dataset commonly used for evaluating depth and egomotion estimates. For depth evaluation, we follow the Eigen train/test split \cite{eigen2015predicting}. For odometry evaluation, we train with sequences \texttt{00}, \texttt{02}, \texttt{05--08}, and test on sequences \texttt{09--10}. Our networks take as input images downsized to 192 $\times$ 640 pixels. 

\subsection{Evaluation Metrics}

To be consistent with existing literature, we evaluate our approach by reporting the standard metrics for each dataset. The mean error is reported in all cases. For KITTI depth evaluation, we follow the usual procedure from \cite{godard2019digging} and report depth accuracy using per-image median ground truth scaling. For odometry evaluation, we report the average translational and rotational errors ($t_{err}(\%)$, $r_{err}(^o/100\text{~m})$) over possible sub-sequences of length (100, 200, $\dots$, 800) metres.
Prior to evaluating these odometry metrics, we align the (unscaled) trajectories with ground truth by applying a constant scale factor to the translation values of the estimated trajectory. 
For evaluation on ScanNet, we report the depth and camera pose metrics as described in \cite{tang2018ba} and, similar to existing literature, use image-wise rescaling to align the predictions with ground truth for both depth and poses.
\subsection{Implementation Details}

We use the same depth and egomotion network structures from \cite{wagstaff2020self2}, implemented in PyTorch \cite{paszke:2017}.\footnote{\href{https://github.com/utiasSTARS/learned_scale_recovery}{github.com/utiasSTARS/learned\_scale\_recovery}.} The encoder-decoder depth network consists of a ResNet18 \cite{he2016deep} encoder and a decoder based on the networks from Monodepth2 \cite{godard2019digging} and DNet \cite{xue2020toward} (we modify Monodepth2 by adding `dense connected prediction' (DCP) layers from DNet). The egomotion network is a seven-layer CNN that takes as input a source-target image pair and returns a 6D pose vector. 
For KITTI and ScanNet experiments, we pretrain our networks on images from the Oxford RobotCar dataset \cite{RobotCarDatasetIJRR}. We train our models on an NVIDIA Titan V GPU for 25, 45, and 15 epochs on the KITTI (odometry and Eigen splits) and ScanNet datasets, respectively, using the Adam optimizer \cite{Kingma:2014} ($\beta_1=0.9,\beta_2=0.999$). We use learning rates of $1\times10^{-4}$ and $2\times10^{-4}$ for the depth and egomotion network, respectively, that are halved four times over the duration of training. During training we apply data augmentation in the form of horizontal flipping and modifications to hue, saturation, contrast, and brightness. 
For the training loss (\cref{eq:loss_terms}), we use $\alpha=0.85$, $\lambda_{P}=1$, $\lambda_{S}=0.05$, and $\lambda_{G}=0.15$. Each sample consists of three consecutive images (the middle target image and its two adjacent source images). We evaluate the loss only at the highest-resolution image scale, rather than using a multi-scale approach. The loss is evaluated in both the forward and inverse directions (in the inverse direction, $\mathbf{I}_t$ is used to reconstruct $\mathbf{I}_s$) to maximize data efficiency. We use the `automasking' and `minimum reprojection' techniques from  \cite{godard2019digging} in addition to the self-discovered mask from \cite{bian2019unsupervised} to ignore/downweight unreliable pixels that break the photometric consistency assumptions. 

\begin{table*}[]
	\centering
	\caption{ScanNet results for the standard two-view test set. The ablation study indicates that more iterations and applying PFT at inference time leads to improved performance. We specify the number of iterations during training ($x$) and inference ($y$) as $x/y$ iter.}
	\label{tab:scannet_results}
	\begin{threeparttable}
		\begin{tabular}{c@{\hspace{0.9\tabcolsep}}c@{\hspace{0.9\tabcolsep}}cccccccc}
			\toprule
			& Method & Abs Rel $\downarrow$ & Sq Rel (m) $\downarrow$ & RMSE (m) $\downarrow$ & RMSE log $\downarrow$ & Sc Inv $\downarrow$ & Rot ($^o$) $\downarrow$ & Tr ($^o$) $\downarrow$ & Tr (cm) $\downarrow$ \\ \midrule
			
			&LSD-SLAM \cite{engel2014lsd} & 0.268 & 0.427 & 0.788 & 0.330 & 0.323 & 4.409 & 34.36 & 21.40 \\ \midrule
			
			\multirow{4}{*}{\textbf{Sup.}}&DeMon \cite{ummenhofer2017demon} & 0.231 & 0.520 & 0.761 & 0.289 & 0.284 & 3.791 & 31.626 & 15.50 \\
			
			&BANet \cite{tang2018ba} & 0.161 & 0.092 & 0.346 & 0.214 & 0.184 & 1.018 & 20.577 & 3.39 \\
			
			&DeepSFM \cite{wei2020deepsfm} & 0.227 & 0.170 & 0.479 & 0.271 & 0.268 & 1.588 & 30.613 & --- \\

			&DeepV2D \cite{teed2019deepv2d} & 0.069 & 0.018 & 0.196 & 0.099 & 0.097 & 0.692 & 11.731 & 1.902 \\

			&DRO (12 iter.) \cite{gu2021dro}  & \textbf{0.053} & \textbf{0.017} & \textbf{0.168} & \textbf{0.081} & \textbf{0.079} & \textbf{0.473} & \textbf{9.219} & \textbf{1.160} \\
			\midrule

			\multirow{2}{*}{\textbf{Self-Sup.}} & DRO (12 iter.) \cite{gu2021dro} & 0.140 & 0.127 & 0.496 & 0.212 & 0.210 & 0.691 & 11.702 & 1.647 \\
			
			&Ours (4/8 iter.) &\textbf{0.088} & \textbf{0.038} & \textbf{0.260} &     \textbf{0.125} &     \textbf{0.122} & \textbf{0.547} & \textbf{10.482} & \textbf{1.290} \\	\midrule
			
			&Ours (using GT Depth) & --- & --- & --- & --- & --- &  0.516 &          10.366 &           1.247 \\  \midrule
			\multirow{8}{*}{\textbf{Ablation}}&1/1 iter. (no PFT) & 0.181 &     0.117 &     0.462 &     0.229 &     0.223 &            1.948 &          43.363 &           5.612 \\
			
			&1/1 iter. & 0.186 &     0.125 &     0.462 &     0.233 &     0.228 &            1.948 &          43.363 &           5.612 \\
			
			& 2/2 iter. (no PFT) & 0.126 &     0.062 &     0.342 &     0.167 &     0.162 &            1.137 &          20.790 & 2.800 \\
			
			& 2/4 iter. (no PFT) & 0.126 &     0.062 &     0.342 &     0.167 &     0.162 & 0.958 &          17.496 &           2.270 \\ 
			
			& 2/4 iter. & 0.111 &     0.054 &     0.302 &     0.150 &     0.147 &            0.821 &          15.195 &           1.952 \\
			
			& 4/4 iter. & 0.092 & 0.040 &  0.266 &     0.129 &     0.126 &   0.695 & 12.340 &  1.669 \\
			
			&4/8 iter. (no PFT) & 0.103 &     0.044 &     0.292 &     0.140 &     0.137 &            0.690 &          12.389 &           1.635 \\\bottomrule
		\end{tabular}
	\end{threeparttable}
\vspace{-4mm}
\end{table*} 

\subsection{Some Notes on PFT}

Our PFT optimization scheme uses the Adam optimizer to update the depth network weights, while the egomotion network weights are held fixed. Following \cite{mccraith2020monocular}, we only update the depth encoder weights instead of updating the full depth network. When minimizing \cref{eq:opt-loss}, we set $\lambda_{prior}=0.1$, and retain the same values for the other coefficients. For every test set sample, 20 optimization `epochs' are performed. 
Instead of returning the network predictions from the final epoch, we average the predictions from the last five epochs to prevent a noisy gradient step from negatively impacting the optimization. 
The runtime for each PFT epoch is 32 ms, which is only moderately larger than the runtime of our four-iteration network (27 ms), and one-iteration network (18 ms).
Lastly, in the course of evaluating our approach on the KITTI odometry dataset, we identified that sample-wise optimization often increases \textit{inter-frame} scale drift, since the PFT procedure can independently shift the scale factor for individual samples within the test sequences. To promote a uniform scale factor across an entire sequence, we incorporate a self-supervised online scale recovery module based on DNet \cite{xue2020toward}; for each image, we estimate the camera height (relative to the ground plane) using our depth prediction and then normalize the corresponding translation prediction using this quantity. We refer the reader to \Cref{tab:vo_opt_ablation} for ablation experiments that show the accuracy of our method without applying this rescaling. It is important to note that this type of scale inconsistency differs from the \textit{inter-network} scale inconsistency that we addressed through the introduction of the tightly-coupled networks. 
\subsection{Evaluation on ScanNet and KITTI}
\subsubsection{ScanNet results} \Cref{tab:scannet_results} presents our results on the ScanNet test split. Our method significantly outperforms the self-supervised variant of DRO and is competitive with supervised methods. We additionally include the pose results from our egomotion network when using the available ground truth depth (instead of our predicted depth) to iteratively warp the source image. The improved pose accuracy indicates that our egomotion network is able to function with an arbitrary depth map that was not present during training. Our ablation study reveals that feedback coupling (through iteration) is crucial for improving accuracy---without this, even the PFT has very little impact. 

\subsubsection{KITTI results} \Cref{tab:kitti_vo_results} and \Cref{fig:topdown_traj_kitti} present the performance of our tightly-coupled system on the KITTI odometry test sequences, and show that our proposed method achieves state-of-the-art egomotion accuracy compared with other learning-based methods. Notably, our coupling approach produces better egomotion estimates than the more simplistic direct coupling methods \cite{ambrus2020two,wang2019recurrent} that treat depth as an input into the egomotion network. 
Our ablation study (\Cref{tab:vo_opt_ablation}) indicates that feedback coupling is crucial for achieving this level of accuracy. Including feedback and applying our PFT strategy  leads to the best performance. \Cref{tab:kitti_depth_results} shows our system's depth accuracy on the Eigen test split; notably, we are competitive with other monocular, self-supervised methods, but lag behind some recent approaches like ManyDepth \cite{watson2021temporal}. We emphasize the utility of our method in producing accurate depth \textit{and} egomotion estimates, whereas ManyDepth and Shu et al. \cite{shu2020feature} are only able to estimate depth accurately.

\begin{table}[]
	\vspace*{1.5mm}
	\caption{KITTI odometry results on the standard test sequences. Methods with test-time PFT are shown in \textbf{bold}.}
	\label{tab:kitti_vo_results}
	\centering
	\begin{threeparttable}
		\begin{tabular}{c@{\hspace{2.5\tabcolsep}}c@{\hspace{2\tabcolsep}}c@{\hspace{2\tabcolsep}}c@{\hspace{2\tabcolsep}}c@{\hspace{2\tabcolsep}}}
			\toprule
			Method & \multicolumn{2}{c}{Seq. \texttt{09}} & \multicolumn{2}{c}{Seq. \texttt{10}} \\ \midrule
			& \begin{tabular}[c]{@{}c@{}}$t_{err}$ \\ $(\%)$ \end{tabular} & \begin{tabular}[c]{@{}c@{}}$r_{err}$ \\ ($^{\circ}$/100 m) \end{tabular} &  \begin{tabular}[c]{@{}c@{}}$t_{err}$ \\$(\%)$ \end{tabular} & \begin{tabular}[c]{@{}c@{}} $r_{err}$ \\ ($^{\circ}$/100 m) \end{tabular}\\ \midrule

			SfMLearner \cite{zhou:2017} & 11.32 & 4.07 & 15.25 & 4.06 \\
			
			SC-SfMLearner \cite{bian2019unsupervised}  & 8.24 & 2.19 & 10.7 & 4.58 \\
			
			Ambrus et al. \cite{ambrus2020two} & 6.72 & 1.69 & 9.52 & 1.59 \\
			
			ManyDepth$^1$ \cite{watson2021temporal}  & 13.36 & 2.93 & 10.18 & 4.25 \\
			
			Wang et al. \cite{wang2019recurrent} & 9.30 & 3.50 & 7.21 & 3.90 \\

			Zou et al. \cite{zou2020learning} & 3.49 & 1.00 & 11.80 & 1.80 \\
			
			\textbf{Shu et al. \cite{shu2020feature}} & 8.75 & 2.11 & 10.67 & 4.91 \\
			
			\textbf{Li et al. \cite{li2020self}} & 5.89 & 3.34 & 4.79 & 0.83 \\
			
			\textbf{DOC \cite{zhang2021deep}}  & 2.02 & 0.61 & 2.29 & 1.10 \\
			
			\textbf{Ours (4-iter)} & \textbf{1.19} & \textbf{0.30} & \textbf{1.34} & \textbf{0.37} \\ \bottomrule
		\end{tabular}
		\begin{tablenotes}
			\item[1] Based on their publicly available model.
		\end{tablenotes}
	\end{threeparttable}
	\vspace{-0.5mm}
\end{table}
\begin{figure*}[]
	\centering
	\begin{subfigure}[]{0.32\textwidth}
		\includegraphics[width=\columnwidth]{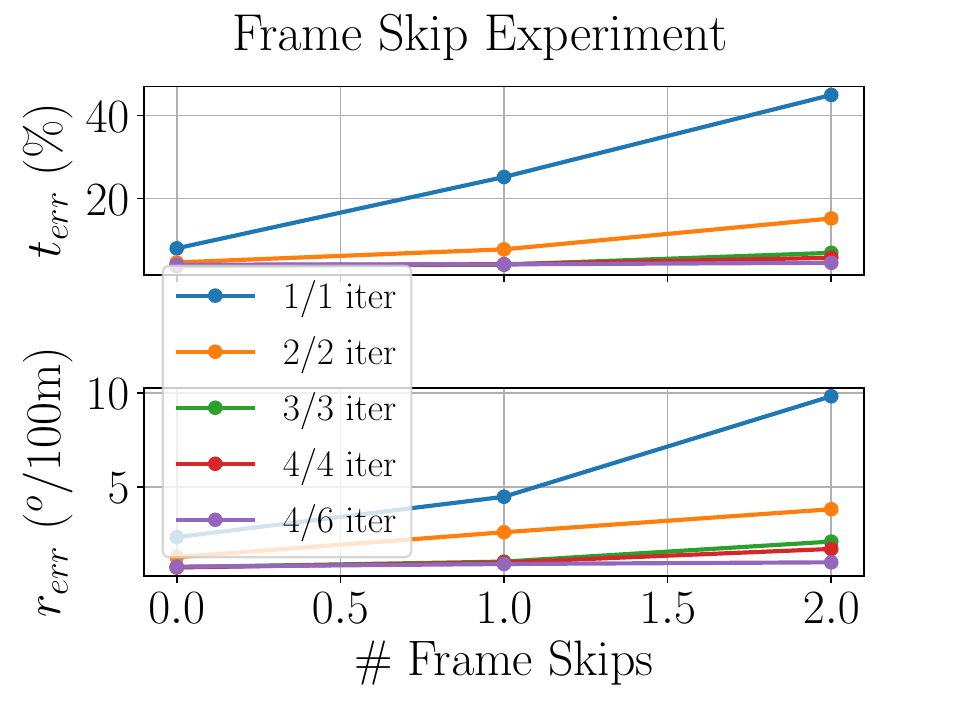}
		\vspace*{-6mm}
		\caption{}
		\label{fig:frame_skip_exp}
	\end{subfigure}\hfill	
	\begin{subfigure}[]{0.32\textwidth}
		\includegraphics[width=\columnwidth]{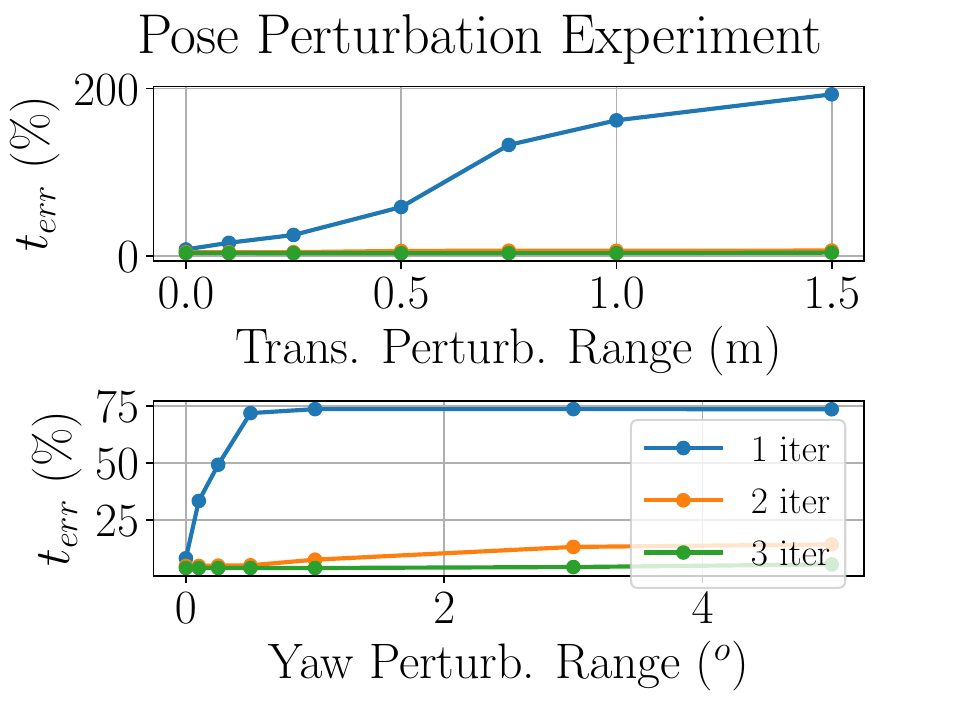}
		\vspace*{-6mm}
		\caption{}
		\label{fig:pose_perturb_exp}
	\end{subfigure}\hfill
	\begin{subfigure}[]{0.32\textwidth}
		\includegraphics[width=\columnwidth]{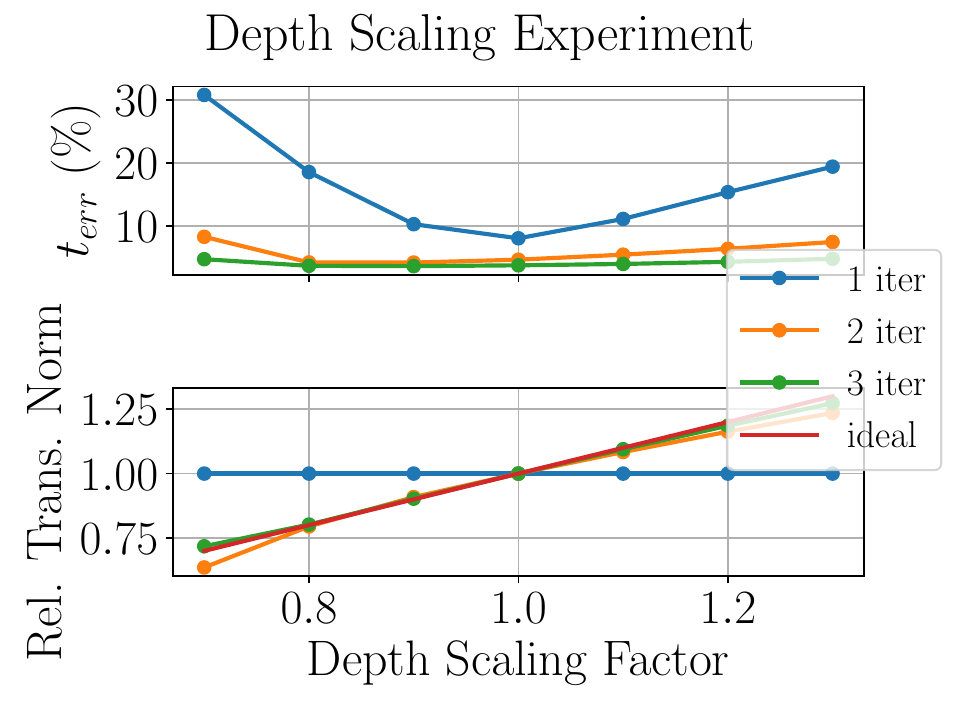}
		\vspace*{-6mm}
		\caption{}
		\label{fig:depth_scaling_exp}	
	\end{subfigure}
	\caption{Summary of the experiments demonstrating the utility of using an iterative egomotion network. Benefits include: (a) improved generalization to unseen data, (b) robustness to initial prediction error, and (c) improved inter-network scale consistency.}
	\label{fig:iterative_experiment_results}
	\vspace{-3mm}
\end{figure*}
\begin{table}[]
	\captionof{table}{Cross-dataset evaluation results on the Oxford RobotCar dataset (sequence \texttt{2014-11-18-13-20-12}).
	}
	\label{tab:oxford_results}
\centering
\begin{threeparttable}
		\begin{tabular}{c@{\hspace{0.95\tabcolsep}}c@{\hspace{0.95\tabcolsep}}c@{\hspace{0.99\tabcolsep}}c@{\hspace{0.95\tabcolsep}}c@{\hspace{0.95\tabcolsep}}}
			\toprule
			 Train. Dataset & \multicolumn{2}{c}{Subseq. \texttt{0}} & \multicolumn{2}{c}{Subseq. \texttt{1}} \\ \midrule
			& $t_{err}$ $(\%)$ & $r_{err}$ ($^{\circ}$/100 m) &  $t_{err}$ $(\%)$  & $r_{err}$ ($^{\circ}$/100 m) \\ \midrule
			Ox. (1-iter) & 7.75 & 3.26 & 9.50 & 3.43 \\
			KITTI (1-iter)  & 36.45  & 12.05  & 32.99  & 13.81  \\
			Rel. Change & -370.32\% & -269.63\% &  -247.26\% & -302.62\% \\ \midrule
			Ox. (3-iter) & 4.84 & 2.18 & 6.46 & 2.38 \\
			KITTI (3-iter) & 5.83  &  3.11 & 11.44 & 4.54  \\
			Rel. change & -20.45\% & -42.66\% &  -77.09\% & -90.76\% \\
			\bottomrule	
		\end{tabular}
	\end{threeparttable}
	\vspace{-4mm}
\end{table}	

\begin{table}[b]
	\centering
	\caption{Ablation study on the KITTI dataset showing the effect of feedback coupling and indirect coupling (through PFT) at inference time. Including both leads to the lowest error. }
	\label{tab:vo_opt_ablation}
	\begin{threeparttable}
		\begin{tabular}{c@{\hspace{0.8\tabcolsep}}c@{\hspace{0.8\tabcolsep}}c@{\hspace{0.8\tabcolsep}}c@{\hspace{0.9\tabcolsep}}c@{\hspace{0.9\tabcolsep}}c@{\hspace{0.9\tabcolsep}}}
			\toprule
			\begin{tabular}[c]{@{}c@{}} \# Egomotion Iter. \\ (train/test) \end{tabular} & PFT Params & \multicolumn{2}{c}{Seq. \texttt{09}} & \multicolumn{2}{c}{Seq. \texttt{10}} \\ \midrule
			& & \begin{tabular}[c]{@{}c@{}} $t_{err}$ \\ $(\%)$ \end{tabular} & \begin{tabular}[c]{@{}c@{}} $r_{err}$ \\ ($^{\circ}$/100 m) \end{tabular} & \begin{tabular}[c]{@{}c@{}} $t_{err}$ \\ $(\%)$ \end{tabular} & \begin{tabular}[c]{@{}c@{}} $r_{err}$ \\ ($^{\circ}$/100 m) \end{tabular} \\ \midrule
			1 / 1 & --- & 8.42 & 2.52 & 9.10 & 4.11  \\
			1 / 3 & --- &  6.16  & 1.51 & 10.13 & 2.74  \\
			2 / 2 & --- & 4.49 & 1.22 & 6.86 & 2.09 \\
			4 / 4 & --- & 2.98 & 0.66 & 3.38 & 1.02 \\
			4 / 6 & --- & 3.02 & 0.69 & 3.03 & 0.93 \\ \midrule
			1 / 1 & $\bm{\theta}_D+\bm{\theta}_E$ & 2.00 & 0.51 & 2.36 & 0.95 \\
			4 / 4 & $\bm{\theta}_D$ & 1.19 & 0.30 & 1.34 & 0.37 \\
			4 / 4 (no DNet scaling) & $\bm{\theta}_D$ & 2.36 & 0.30 & 2.36 & 0.38 \\
			4 / 4 (no $\mathcal{L}_{prior}$) & $\bm{\theta}_D$ & 1.79 & 0.275 & 2.13 & 0.46 \\
			4 / 4 & $\bm{\theta}_D+\bm{\theta}_E$ & 1.28 & 0.31 & 1.36 & 0.35 \\
			\bottomrule
		\end{tabular}
	\end{threeparttable}
\end{table}

\subsubsection{Feedback coupling experiments}
\label{sec:zeroth_order_optimizer}

Lastly, we present evidence that supports the value of using feedback for egomotion estimation. First, we evaluate how feedback during iteration improves generalization to unseen data. 
We do so by evaluating our network on image pairs with inter-frame perspective changes that are significantly larger than those within the dataset (which can occur due to increased camera velocity or decreased camera frame rate). 
We modify the KITTI dataset by skipping images (i.e., adopting a stride greater than one) and then evaluate test sequence odometry error as a function of the increased perspective change. \Cref{fig:frame_skip_exp} shows that as the number of iterations increases, the generalization performance improves. Second, we include a cross-dataset evaluation of our KITTI-trained model on the Oxford RobotCar dataset \cite{RobotCarDatasetIJRR}. \Cref{tab:oxford_results} depicts these results, which indicate that iteration is crucial for generalizing to unseen data. 

Next, we verify how robust the feedback coupling mechanism is by demonstrating the ability to minimize error when large perturbations are applied to the initial egomotion prediction. To investigate this aspect, we apply perturbations (in the forward translation direction, and along the yaw axis) to all of the egomotion predictions within our KITTI test sequences and then evaluate the overall odometry error. We repeat this experiment with increasing perturbation ranges, and summarize the results in \Cref{fig:pose_perturb_exp}. The increase in error from adding perturbations can be completely mitigated by applying only two extra iterations. During these subsequent iterations, the network, which takes as input the (erroneously) warped source image, produces corrections that effectively compensate for the initial perturbation. 

Finally, we verify our claim that appropriate coupling of the network predictions improves inter-network scale consistency. For this experiment, we rescale the depth predictions for the KITTI test sequences by a constant scale factor and then observe how the scale of the translation norm of the egomotion predictions changes in response to the modified depth. \Cref{fig:depth_scaling_exp} illustrates that the average change in the translation norms (for the iterative models) are proportional to the applied depth scaling factor. This provides support for the notion that the egomotion network is able to infer scale from depth predictions via our feedback coupling strategy.
\begin{table*}[]
	\centering
	\caption{Monocular depth prediction results for self-supervised methods on the Eigen test split \cite{eigen2015predicting}. We report results from our depth network following inference time PFT. In all other cases, we report the results from the image resolution closest to ours. }
	\label{tab:kitti_depth_results}
	\begin{threeparttable}
		\begin{tabular}{ccccccccccc}
			\toprule

			Method & Test-Time PFT & Abs Rel $\downarrow$ & Sq Rel (m) $\downarrow$ & RMSE (m) $\downarrow$ & RMSE log $\downarrow$ & $\delta < 1.25$ $\uparrow$ & $\delta < 1.25^2$ $\uparrow$ & $\delta < 1.25^3$$\uparrow$ \\ \midrule
			
			SC-SfMLearner \cite{bian2019unsupervised} && 0.128 & 1.047 & 5.234 & 0.208 & 0.846 & 0.947 & 0.976 \\
			
			Nabavi et al. \cite{Nabavi2020iter} && 0.160 & 1.195 & 5.916 & 0.245 & 0.774 & 0.917 & 0.964 \\
			
			MonoDepth2 \cite{godard2019digging} && 0.115 & 0.903 & 4.863 & 0.193 & 0.877 & 0.959 & 0.981 \\
			
			PackNet-SfM \cite{guizilini20203d} && 0.107 & 0.803 & 4.566 & 0.197 & 0.876 & 0.957 & 0.980 \\
			
			Guizilini et al. \cite{guizilini2020robust} && 0.111 & 0.785 & 4.601 & 0.189 & 0.878 & --- & --- \\
			
			DRO \cite{gu2021dro} (12 iter.) && \textbf{0.088} & 0.797 & 4.464 & 0.212 & 0.899 & 0.959 & 0.980 \\ 
			
			GLNet \cite{chen2019self} & \checkmark & 0.099 & 0.796 & 4.743 & 0.186 & 0.884 & 0.955 & 0.979 \\
			
			ManyDepth \cite{watson2021temporal}  & \checkmark & 0.090 & 0.713 & 4.261 & 0.170 & 0.914 & \textbf{0.966} & \textbf{0.983} \\
			
			McCraith et al. \cite{mccraith2020monocular} & \checkmark & 0.089 & 0.747 & 4.275 & 0.173 & 0.912 & 0.964 & 0.982 \\
			
			Shu et al. \cite{shu2020feature} & \checkmark & \textbf{0.088} & \textbf{0.712} & \textbf{4.137} & \textbf{0.169} & \textbf{0.915} & 0.965 & 0.982 \\
			
			CoMoDA \cite{kuznietsov2021comoda} & \checkmark & 0.102 & 0.871 & 4.596 & 0.183 & 0.898 & 0.961 & 0.981 \\
			
			Ours (4-iter) & \checkmark & 0.097 & 0.791 & 4.383 & 0.178 & 0.896 & 0.961 & 0.982 \\			
			
			\bottomrule
		\end{tabular}
	\end{threeparttable}
	\vspace{-4mm}
\end{table*}
\section{Conclusion}

In this paper, we demonstrated that an appropriate coupling of depth and egomotion networks improves performance in self-supervised SfM. We introduced a taxonomy of coupling methods and discussed the potential benefits of incorporating each method to promote consistency between depth and egomotion predictions at training and at inference time. We noted the particular importance of feedback coupling as a method to iteratively update an initial prediction to minimize the overall loss function. Building on these insights, we presented our own \textit{tightly-coupled} approach and, through extensive experiments, showed that our system is consistently accurate on indoor and outdoor datasets and achieves state-of-the-art accuracy on several key benchmarks. As future work, we aim to generalize our tightly-coupled framework to incorporate additional modalities such as inertial data.
\vspace{-0.25cm}
\section*{Acknowledgments}
We gratefully acknowledge the contribution of NVIDIA Corporation, who provided the Titan V GPU used for this research through their Hardware Grant Program.
%
\vspace{-0.2cm}
\bibliographystyle{IEEEcaps}
\bibliography{references.bib}
\end{document}